\documentclass[runningheads]{llncs}
\usepackage{hyperref}

\usepackage{graphicx}
\usepackage{algorithm}
\usepackage{algorithmic}
\usepackage{adjustbox}
\usepackage{booktabs}
\usepackage{makecell}
\usepackage{amsmath}
\begin{document}

\title{ReCCoVER: Detecting Causal Confusion for Explainable Reinforcement Learning}

\titlerunning{ReCCoVER: Detecting Causal Confusion for Explainable RL}
%
\author{Jasmina Gajcin\inst{1}\orcidID{0000-0002-8731-1236
} \and
Ivana Dusparic\inst{1}\orcidID{0000-0003-0621-5400}}

\authorrunning{}
%
\institute{Trinity College Dublin, Dublin, Ireland}
\maketitle  
\begin{abstract}

Despite notable results in various fields over the recent years, deep reinforcement learning (DRL) algorithms lack transparency, affecting user trust and hindering their deployment to high-risk tasks. Causal confusion refers to a phenomenon where an agent learns spurious correlations between features which might not hold across the entire state space, preventing safe deployment to real tasks where such correlations might be broken. In this work, we examine whether an agent relies on spurious correlations in critical states, and propose an alternative subset of features on which it should base its decisions instead, to make it less susceptible to causal confusion. Our goal is to increase transparency of DRL agents by exposing the influence of learned spurious correlations on its decisions, and offering advice to developers about feature selection in different parts of state space, to avoid causal confusion. We propose ReCCoVER, an algorithm which detects causal confusion in agent's reasoning before deployment, by executing its policy in alternative environments where certain correlations between features do not hold. We demonstrate our approach in the taxi and grid world environments, where ReCCoVER detects states in which an agent relies on spurious correlations and offers a set of features that should be considered instead.

\end{abstract}
\section{Introduction}

Understanding decisions of RL agents can increase users' trust and encourage collaboration with the system, prevent discrimination or uncover surprising behavior \cite{puiutta2020explainable}. One of the main tasks of explainability in RL is, however, to ensure that agent's behavior is correct and not prone to mistakes. This is a necessary step before agent can be deployed to a real-life task. 

\begin{figure}[t]
    \centering
    \includegraphics[scale=0.4]{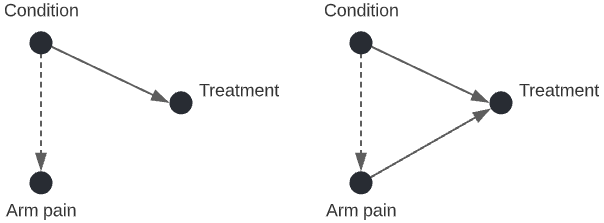}
    \caption{Different causal structures in different situations for a medical decision system. Agent uses features $Condition$ and $Arm pain$ to decide on $Treatment$. Dotted lines represent causal relationships between features, while full lines show whether agent should rely on the feature for its decision. Left: if $Condition = Heart Attack$, arm pain should be disregarded. Right: If $Condition = Arm Injury$, arm pain is an important factor in determining treatment.}
    \label{fig1}
\end{figure}

In this work, we focus on one specific obstacle to successful deployment of RL agents -- \textit{causal confusion}. Causal confusion is a phenomenon which occurs when agent learns to rely on spurious correlations between the features which might not hold over the entire state space. For illustration, we refer to an example of a medical decision-making system for assigning urgent care to patients \cite{lyle2021resolving}, which learns to rely on the correlation between arm pain and heart attacks, resulting in system suggesting emergency care for a patient with a minor arm injury. In other words, the system failed to learn the causal structure of the task -- that heart attack is the true reason for administering urgent care and arm pain is just a side effect that is highly correlated, but should not affect the decision (Figure \ref{fig1}). Reliance on spurious correlations becomes an issue when agent is confronted with a situation in which the correlation does not hold. For this reason, it is necessary to verify that agent's behavior does not depend on any spurious correlations before deployment to a real-life task, even if agent performs optimally in experimental setting. Causal confusion was first detected in imitation learning (IL), where RL agent can learn to rely on spurious correlations in expert's observations which do not hold in the RL environment \cite{de2019causal}. Causal confusion was further examined in RL and, to mitigate its effects, a method for targeted exploration was proposed \cite{lyle2021resolving}, forcing the agent to visit states that challenge potential learned spurious correlations.

In this work, we address the problem of causal confusion in RL from the perspective of explainability, and aim to increase understanding of agent's behavior in critical states, by examining whether agent relies on spurious correlations to make a decision. To achieve the above, we propose ReCCoVER (\underline{Re}cognizing \underline{C}ausal \underline{Co}nfusion for \underline{V}erifiable and \underline{E}xplainable \underline{R}L), an approach for detecting and correcting situations in which an agent relies on damaging spurious correlations. If spurious correlation is uncovered, we identify the area of state space in which relying on such correlation is damaging and ReCCoVER generates an alternative subset of features that the agent should instead focus on in that state subspace, to avoid causal confusion. It is important to note that the subset of features agent should rely on can differ between states -- a feature can be useful in one state, but might need to be ignored in another (Figure \ref{fig1}). 
To make our explanations concise, we verify agent's reasoning only in critical states. Specifically, we focus on states in which agent reaches local maximum in terms of state-value function. Since agent has an opportunity to achieve a large reward or complete a subtask in local maxima states \cite{sequeira2020interestingness}, these decisions are some of the most important in agent's execution.
To uncover the possibility of spurious correlation in state $s$, we make use of the fact that a policy relying on such correlation will fail to generalize to situations where the correlation is broken. To that end,  we create \textit{alternative environments} by performing causal interventions on features and explore the behavior of the policy in different settings. Taking the example of the medical system described above, we explore what makes it decide on urgent care for a patient with heart attack by forcing arm pain to be absent and observing how well agent performs in the alternative situation. To find out whether a policy relying on a different set of features would generalize better in the alternative environment, we propose a method for training a feature-parametrized policy, which simultaneously learns a separate policy for each possible subset of features. Performance of this policy in alternative environments is used to detect whether ignoring certain features during training could prevent spurious correlations from being learned and produce correct behavior in an alternative environment in which agent's original policy fails. This way we can make sure that agents make decisions in critical states based on the correct features (e.g. suggesting urgent care because of the heart-attack and not arm pain), and do not rely on damaging correlations. 

Current work in causal confusion focuses on learning a robust policy that does not rely on spurious correlations \cite{lyle2021resolving}. However, current approaches can only detect spurious correlations in agent's reasoning after recording a drop in performance when the agent is deployed to an environment where such correlations do not hold. In contrast, we focus on verifying the behavior of a trained policy before deployment, to ensure that it does not rely on damaging spurious correlations, and propose improvements to make policy less prone to causal confusion and thus safer for deployment. Most similar to our work is the original work on causal confusion, which proposes 
a method for uncovering the true causes of expert's actions when an RL policy is learned using IL \cite{de2019causal}. However, the work assumes that the true causes are constant throughout the episode, and that the same set of features should be considered and same features should be ignored in each state. Additionally, the work detects causal confusion which originates as a consequence of distributional shift between expert's observations and agent's environment. In contrast, we observe that a single feature can be both useful and confusing, depending on the situation -- arm pain should not be a cause of a decision if a patient has a heart attack, but should be considered if they suffer from an arm injury. Moreover, we focus on a RL setting where causal confusion stems from agent learning spurious correlations throughout its experience in the environment, and create alternative environments in which such correlations do not hold to verify agent's reasoning.

By proposing ReCCoVER, the main contributions of this paper are:
\begin{enumerate}
    \item A method for extracting critical states in which agent's behavior needs to be verified. For each critical state, ReCCoVER generates a set of alternative environments in which certain correlations between features are broken, to simulate conditions that the agent did not observe often during training.
    \item An approach for detecting spurious correlations in agent's policy by examining agent's performance in alternative environments where learned correlations do not hold. If spurious correlation is detected, we identify the area of state space in which relying on that correlation is damaging and ReCCoVER proposes a subset of features that agent should rely on in that area instead. 
\end{enumerate}

We test and evaluate ReCCoVER in the taxi and minigrid traffic environments, where we uncover parts of the state space where causal confusion is damaging and propose a different subset of features agent should rely on, to avoid learning spurious correlations. Full ReCCoVER code and evaluation environments are available at \url{https://github.com/anonymous902109/reccover}.

\section{Related Work}

In this section we present a short description of structural causal models, along with a brief overview of current methods in explainable RL. 

\subsection{Structural Causal Models (SCM)}

Structural causal models \cite{scm} represent causal relationships between variables and consist of two main parts: a directed acyclic graph (DAG), and a set of structural equations. Each node in the DAG is associated with either an observable \((\{X_i\}^n_{i=1})\) or an unobservable variable \((\{U_i\}^{m}_{i=1})\), and presence of a directed edge between two nodes implies a causal relationship. Structural equations can be assigned to edges in order to describe the strength of the causal effect:

\begin{equation}
    X_i = f_i(PA_i, U_i)
\end{equation}

In other words, value of each observable variable \(X_i\) is directed by the values of its parent variables \(PA_i\) and unobserved variables through some function \(f_i\).

Causal interventions are operations on the SCM that help us detect a cause-effect relationship between two variables. Intervention on variable $A$ is usually denoted by the do-operator: $do(A = v)$, and represents setting value of $A$ to $v$ \cite{peters2017elements}. Intervening on a variable removes any incoming causal arcs to that variable, because it forcibly sets the value of the variable, regardless of the other variables that might be causing it. Outgoing causal relationships will, however, remain intact. When intervening on features of an RL state, we use the $do(f \rightarrow v; s)$ notation to describe forcibly setting feature $f$ to $v$ in state $s$.

\subsection{Explainable Reinforcement Learning (XRL)}

Recent years have seen a rise in developing methods for explaining RL agents \cite{puiutta2020explainable}. According to their scope, XRL approaches can be divided into \textit{local} and \textit{global}. While local XRL methods explain one decision \cite{olson2019counterfactual,madumal2020explainable,van2018contrastive}, the aim of global explanations is to understand the behavior of a policy as a whole \cite{amir2018highlights,coppens2019distilling,sequeira2020interestingness,verma2018programmatically,gajcin2021contrastive}. Within this classification, ReCCoVER can be considered a hybrid method -- while we aim to verify policy behavior locally, in specific, critical states, we also provide a way to extract critical states automatically, enabling global explanations to be generated without the need for user input. 

Feature attribution is a local method for explaining a decision by assigning importance to input features, depending on their contribution to the output \cite{burkart2021survey}. Feature attribution is often referred to as feature importance \cite{carvalho2019machine}. In supervised learning, feature attribution is commonly used for explaining and verifying behavior of black-box models. For example, LIME \cite{lundberg2017unified} generates an intrinsically interpretable local model which approximates behavior of the black-box system around the instance in question. Feature importance can be then extracted from the interpretable model, for example as feature weights in case of linear LIME. Similarly, saliency maps are used in image-based supervised learning tasks to highlight parts of the image which contributed the most to the decision \cite{simonyan2013deep}. 

In XRL, feature attribution methods such as saliency maps and LIME have been adopted from supervised learning to explain decisions of RL agents \cite{greydanus2018visualizing,puri2019explain,dethise2019cracking}. Similarly, the aim of this work is to explain agent's behavior by examining the influence of individual features on the decision. However, instead of only showing which features contributed to the decision, we explore whether an agent is relying on spurious correlations in that state, and if so, offer an alternative set of features that agent should instead be considering. This enables the developer to not only detect errors in agent's reasoning, but also correct them.

\section{ReCCoVER}
\label{Causal confusion}

In this section, we posit a method for examining critical states and detecting potential spurious correlations in agent's decision-making process, which we call ReCCoVER :\underline{Re}covering \underline{C}ausal \underline{Co}nfusion for \underline{V}erifiable and \underline{E}xplainable \underline{R}einforcement Learning. As input, ReCCoVER requires agent's learnt policy $\pi$, state-value function $V_\pi(s)$, feature set $\mathcal{F}$ and a set of possible feature subsets $\mathcal{G}$. If the entire feature space is searched, $\mathcal{G}$ is the power set of $\mathcal{F}$, $\mathcal{G} = \mathcal{P}(\mathcal{F})$.

\begin{figure}[t]
    \centering
    \includegraphics[width=\textwidth]{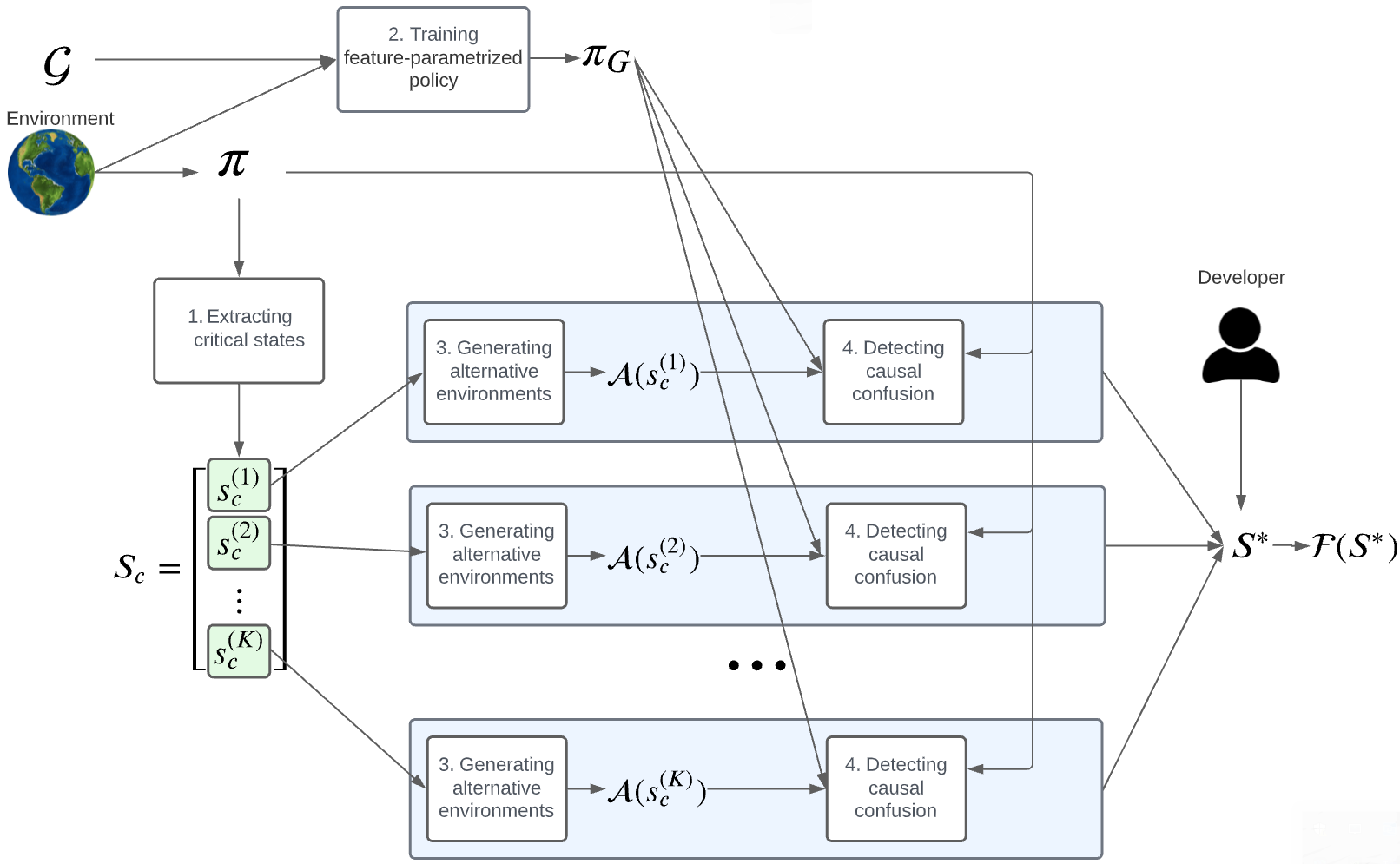}
    \caption{Overview of ReCCoVER: to explain decisions of policy $\pi$, a set of critical states $S_c$ is extracted. Feature-parametrized policy $\pi_G$ is trained to simultaneously learn a separate policy for each subset of features in $\mathcal{G}$. For each critical state $s_c^{(i)}$, a set of alternative environments $\mathcal{A}(s_c^{(i)})$ is generated, and performance of policy $\pi$ and policies corresponding to different feature subsets in $\pi_G$ are evaluated in $\mathcal{A}(s_c^{(i)})$. Alternative environments in which causal confusion is detected are manually examined to extract an area of subspace $S^*$ where relying on causal confusion damages performance of $\pi$ and a different subset of features $\mathcal{F}(S^*)$ is proposed to be used in states in $S^*$.}
    \label{approach}
\end{figure}

We start by describing an approach for gathering critical states in agent's execution (Section \ref{Extracting critical states}). Since actions in these states can bring large reward, we focus on verifying agent's behavior in these states. We then describe the process of simultaneously learning a separate policy for each subset of features in $\mathcal{G}$ (Section \ref{Training graph-parametrized policy}). We propose a method for generating alternative environments by performing interventions on state features (Section \ref{Generating alternative worlds}). Since setting a value of a feature can break its correlation with other features, we use policy performance in alternative environments to observe how policies depending on different features handle different correlation settings. Finally, we describe how causal confusion can be detected in alternative environments by executing a feature-parametrized policy and observing whether ignoring specific features during training prevents the agent from learning spurious correlations (Section \ref{Detecting causal confusion}). An overview of the ReCCoVER approach is shown in Figure \ref{approach} and Algorithm 2.

\begin{algorithm}[ht]
\textbf{Input:} Policy $\pi$, state-value function $V_\pi$, feature set $\mathcal{F}$, feature subsets $\mathcal{G}$\\
\textbf{Parameters:} $\alpha$, $k$\\
\textbf{Output:} Feature subset $\mathcal{F}(S^*)$ 
    \begin{algorithmic}[1]
        \caption{ReCCoVER algorithm}
        
        \STATE Extract critical states $S_c$
        \STATE Train a feature-parametrized policy $\pi_G$
        
        \FOR{$s_c \in S_c$} {
            \STATE Generate alternative environments $\mathcal{A}(s_c)$ by intervening on each feature in $\mathcal{F}$
            \STATE Filter $\mathcal{A}(s_c)$ based on state novelty
            
            \FOR{$A \in \mathcal{A}(s_c)$}{
                \STATE Evaluate policy $\pi$ in $A$ for $k$ steps and record return $R_\pi$
                
                \FOR{$G_i \in \mathcal{G}$}{
                    \STATE Evaluate policy $\pi_G(G_i)$ in $A$ for $k$ steps and record return $R(G_i)$ 
                }\ENDFOR
                \IF{$R_\pi \ll R(G_j)$ for some $G_j \in \mathcal{G}$}{  
                    \STATE Identify subspace $S^* \subseteq S$ where detected spurious correlation is damaging
                    \STATE Propose feature subset $G_j$ to be used in states in $S^*$
                }\ENDIF
            }\ENDFOR
        }\ENDFOR
    \end{algorithmic}
\end{algorithm}

\subsection{Extracting Critical States}
\label{Extracting critical states}

Naturally, agent's behavior must be verified as a whole before deployment to a real-life task. However, it can be costly to explain agent's behavior in each state. Additionally, making a sub-optimal action is riskier in certain states than in others. For example, taking a slightly longer route is less dangerous for a self-driving car than incorrectly reacting in a near-collision situation. Thus, we focus on ensuring that agent behaves correctly in selected, critical states. 

Inspired by the idea of \textit{interesting states} \cite{sequeira2020interestingness}, we consider critical states to be those in which an agent reaches a local maximum. Formally, given policy $\pi$ and its state-value function $V_\pi(s)$, a set of local maxima states $S_{c}$ is defined as:

\begin{equation}
        S_{c} = \{s \in S | V_\pi(s) \geq V_\pi(s'), \forall s' \in T_s \}
\end{equation}

where $T_s$ is a set of states that agent can transition to from $s$ by taking any available action $a \in \mathcal{A}$. Local maxima states represent situations that are preferable for the agent, and often correspond to a larger reward, due to achieving a sub goal or finishing the task. Since making the right decision in these states can yield a large reward, it is important to ensure agent's reasoning is correct and does not rely on spurious correlations in these states. The output of this part of the method is a set of critical states $S_{c}$.

\subsection{Training Feature-Parametrized Policy}
\label{Training graph-parametrized policy}

\begin{figure}[t]
    \centering
    \includegraphics[scale=0.4]{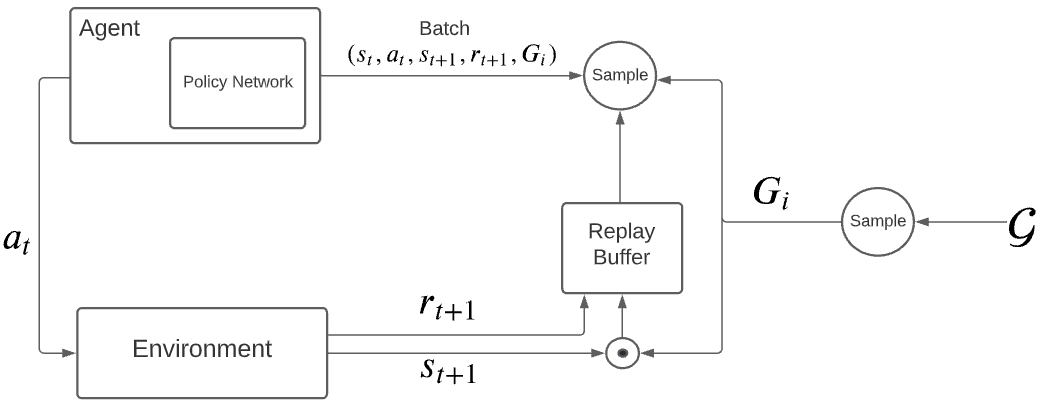}
    \caption{Feature-parametrized policy training: each episode \(i\), a random feature subset \(G_i\) is generated and used to mask agent states. At the same time, \(G_i\) is used as a parameter for the sampling function, to ensure that only policy relying on $G_i$ is updated.}
    \label{fig-flow-chart}
\end{figure}

Knowing a separate policy for each feature subset helps us understand how agent's behavior changes when its access to certain features is denied during training. 
To train a separate policy for each subset of features, we adapt the approach proposed by \cite{de2019causal} from the IL to an RL setting. To uncover the subset of features an agent should rely on in order to avoid causal confusion, authors in  \cite{de2019causal} propose an approach for training a  feature-parametrized neural network to simultaneously learn a separate policy for each possible subset of features. Each feature subset is encoded as a binary vector, with each coordinate indicating whether the specific feature is present. During training, feature subsets are iteratively sampled and only policy relying on the chosen subset of features is updated.

We adapt this approach from IL to DRL setting. We also represent each feature subset as a binary vector, where each coordinate denotes presence of a feature. However, instead of supervised methods used in the original work, we use DQN \cite{mnih2013playing} to learn a policy. In every iteration of the algorithm, a binary vector is sampled at random and the feature-parametrized policy is updated based only on features present in the chosen vector. In this way, multiple policies that rely only on features available in their corresponding vectors are learned simultaneously. High-level overview of the training process for the feature-parametrized DRL policy is shown in Figure \ref{fig-flow-chart} and further laid out in Algorithm 1. The output of this stage of ReCCoVER is a feature-parametrized policy $\pi_G$.

\begin{algorithm}[!ht]
\textbf{Input:} Set of feature subsets $\mathcal{G}$ \\
\textbf{Output:} Feature-parametrized policy $\pi_G$
    \begin{algorithmic}[1]
        \STATE Randomly initialize appropriate network(s)
        \STATE Initialize replay memory buffer \(\mathcal{D}\)
        \WHILE{loss not converged} {
            \STATE Receive initial observation \(s_{0}\)
            \STATE Sample random feature subset \(G_{i} \in \mathcal{G}\)
            \WHILE{episode not over}{
                \STATE Mask current state \(s_t^* = [s_{t} \odot G_{i}; G_{i}]\)
                \STATE Select action \(a_t\) based on policy network 
                \STATE Execute \(a_t\), receive reward \(r_{t+1}\) and new state \(s_{t+1}\)
                \STATE Update state \(s_t = s_{t + 1}\)
                \STATE Mask new state \(s_{t+1}^* = [s_{t+1}  \odot G_{i}; G_{i}]\)
                \STATE Store \((s_t^*, a_t, r_{t+1}, s_{t+1}^*, G_{i})\) in \(\mathcal{D}\)
                \STATE Sample batch from \(\mathcal{D}\) with feauture subset \(G_i\)
                \STATE 
                Calculate the loss specific to the DRL algorithm 
                \STATE Perform gradient descent update based on the loss
            }\ENDWHILE
        } \ENDWHILE
    \end{algorithmic}
    \caption{Training a feature-parametrized policy $\pi_G$}
\end{algorithm}

\subsection{Generating Alternative Environments}
\label{Generating alternative worlds}

A policy that relies on a damaging spurious correlation will fail to generalize to an environment where such correlation does not hold. To test whether agent learned such correlations, we generate alternative environments in which specific correlations between features do not hold and evaluate the policy in them. To break correlation between two features we use causal interventions. Specifically, for two features $A$ and $B$, where $A$ causes $B$ ($A \rightarrow B$), performing an intervention on $B$ and setting it to a fixed value cancels any correlation between the two. 

To examine how well a policy handles a change in correlations between features in a critical state $s_c$, we generate a set of alternative environments $\mathcal{A}(s_c)$ by intervening on specific state features of $s_c$. Environment $A(f \rightarrow v; s_c)$ denotes an environment with starting state $s_c$, in which feature $f$ is forcibly set to value $v$. By setting a feature to a specific value, we break any correlation between that feature and features that are causing it. 

For a task with $N$ features, where each feature can take $M$ different values, there are $N \cdot M$ possible alternative environments. Seeing how generating such large number of environments might be unfeasible, we evaluate the policy in the selected subset of alternative environments. Since we wish to test agent's behavior in unexpected situations, where learned correlations might be broken, we adopt the idea of \textit{state novelty} \cite{csimcsek2004using}. Intuitively, to explain agent's action in state $s_c$, we only perform intervention $do(f \rightarrow v; s_c)$ on state $s_c$ if it leads to a highly novel state $s'$. We generate a data set of agent's transitions $\mathcal{D}$ by unrolling the policy in the environment and calculate novelty of a state $s$ as:

\begin{equation}
    \mathcal{N}(s) =
    \begin{cases}
    \frac{1}{\sqrt{n(s)}},& \text{if } n(s) \geq 1\\
    1,  & \text{otherwise}
\end{cases}
\end{equation}

where $n(s)$ is the number of occurrences in agent's experience $\mathcal{D}$.

For a critical state $s_c$ we generate a set of alternative environments by performing interventions on $s_c$ that lead to highly novel states:

\begin{equation}
    \mathcal{A}(s_c) = \{A(f \rightarrow v; s_c) | \mathcal{N}(s') > \alpha, s' = do(f \rightarrow v; s_c)\}
\end{equation}

where $\alpha$ is a threshold value, denoting how novel a state must be to be considered. The output from this stage of ReCCoVER is a set of alternative environments $\mathcal{A}(s_c)$ for each critical state $s_c \in S_c$.

\subsection{Detecting Causal Confusion}
\label{Detecting causal confusion}

Given a set of alternative environments $\mathcal{A}(s_c)$ for a state $s_c$ being explained, we can execute policy $\pi$ to test its robustness in each of them. If $\pi$ fails in environment $A(s_c)$, one explanation for that could be that agent's reliance on a spurious correlation has lead to poor generalization to an alternative environment where such correlation is broken. However, it is also possible that by intervening on a feature, we made the task much more difficult or impossible to complete in the alternative environment (e.g., changed the goal position to a far away or unreachable position). The difference is that, if reliance on a spurious correlation is the cause of failure, then a policy which ignores features that contribute to spurious correlation in that state would perform well, as it would not be able to learn the damaging correlation. On the other hand, if environment $A(s_c)$ is indeed unsolvable, any policy relying on any subset of features will fail as well. 

To examine whether agent's policy $\pi$ relies on spurious correlations in a specific critical state, we compare its performance in alternative environments with the performance of feature-parametrized policy $\pi_G$. By evaluating individual policies of $\pi_G$, we can see how policies relying on different subsets of features differ in behavior in a critical state. Given a set of alternative environments $\mathcal{A}(s_c)$ in critical state $s_c$, agent's policy $\pi$ is executed in each of them for a fixed number of steps $k$. Similarly, individual policies corresponding to different feature subsets from $\pi_G$ are executed in each alternative environment for the same number of steps. In each alternative environment $A(s_c)$, we compare the returns of agent's policy $\pi$, with returns of individual policies in $\pi_G$.

Causal confusion is detected in an alternative environment $A(s_c)$ if agent's policy $\pi$ fails to generalize to $A(s_c)$, but policy $\pi_{G}(G')$ corresponding to subset of features $G'$ shows good performance. This means that $\pi$ learned to rely on a spurious correlation in $s_c$ which $\pi_{G}(G')$ did not learn, as it did not consider certain features. Formally, for feature-parametrized policy $\pi_G$, agent's policy $\pi$ and an alternative environment $A(f \rightarrow v;s_c)$, ReCCoVER detects causal confusion if there exists a policy $\pi_G(G')$ trained on a subset of features $G'$, that performs significantly better than $\pi$ in $A(f \rightarrow v;s_c)$:

\begin{equation}
    \mathcal{R}^k(A(f, v;s_c), \pi) \ll \mathcal{R}^k(A(f, v;s_c), \pi_G(G'))
\end{equation}

where $\mathcal{R}^k(A, \pi)$ is a return achieved in environment $A$ after following policy $\pi$ for $k$ steps. Defining what is considered a ``significantly better performance'' can depend on the task. In this work, we assume RL tasks in which failure is met with a large negative reward, thus allowing for more straightforward identification of poor generalization performance.  On the contrary, if $\pi$ is at least as successful as other policies in $\pi_G$ in all alternative environments, we assume that its behavior is satisfactory and not reliant on spurious correlations.

An environment $A(s_c)$ where causal confusion is detected uncovers a part of the state space in which reliance on spurious correlations damages agent's performance. We expect that alternative environments in which causal confusion is detected will offer an insight into which parts of the environment agent cannot conquer due to its reliance on spurious correlations. As our work is preliminary and of exploratory nature, we manually examine the alternative environments where causal confusion is detected, to extract parts of the state space in which an agent should rely on a modified subset of features. Ideally, this process should be automated, and problematic state subspaces directly extracted. If causal confusion is detected in an alternative environment $A(s_c)$, where $\pi_G(G')$ outperforms $\pi$, and a human expert manually extracts a subset of problematic states $S^* \subseteq S$, we propose that the agent should rely on features from $G'$ in states from $S^*$.

\section{Evaluation Scenarios and Settings}

In this section, we describe two environments in which we evaluate ReCCoVER.

\subsubsection*{Taxi Environment} is modelled after the OpenAI \cite{brockman2016openai} taxi environment. Agent's goal in this task is to navigate a \(5\times5\) grid world, pick up a passenger from a designated location and drop them off at their destination. Each step in the environment yields a \(-1\) penalty, picking up the passenger wins $+10$ reward and for successfully completing the task the agent receives \(+20\) reward. Episodes terminate upon task completion, or after \(200\) time steps. 

In the original environment, the agent has information about the taxi's position, passenger's location and destination. We augment the state space with another feature -- a passenger descriptor, which has a high influence on passenger's choice of destination, in order to create potential for spurious correlations to be learned in the environment. Using the passenger descriptor as a proxy for destination will have no negative consequences as long as the two features are highly correlated. However, if the agent that relies on the descriptor encounters a situation in which the descriptor is not indicative of the destination, it will fail to drop off the passenger. It is imperative from the perspective of generalization and explainability, for the agent to recognize that, despite this correlation, destination is ultimately the important feature as it will always indicate the correct drop-off coordinates, regardless of the descriptor value. In each state, agents should rely only on the original features, and ignore the passenger descriptor.

\begin{figure}[t]
    \centering
    \includegraphics[scale=0.4]{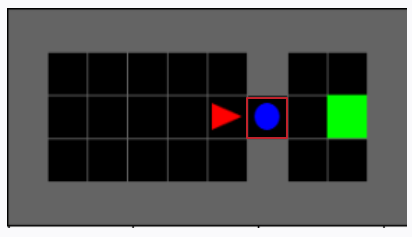}
    \caption{MiniGridworld traffic environment: Agent (red) needs to reach the goal (green), avoid collision with the other vehicle (blue) and obey the traffic light (outlined red).}
    \label{gridworld-traffic}
\end{figure}

\subsubsection*{MiniGrid Traffic Environment} is based on grid world environment \cite{gym_minigrid}. We extend it into a traffic environment by setting up a traffic light, and another vehicle in the environment in front of the agent. Reaching the goal requires obeying the traffic light and avoiding collision with the other vehicle. The vehicle in front of the agent drives ahead whenever possible, while obeying the traffic light. The agent has access to its own location, location of the goal, location and the current action of the other vehicle, traffic light location and light color. The agent can decide at each step whether to move forward or remain in place. Each step has a living $-1$ penalty, while crashing into the other vehicle or violating the red light costs $-10$ and terminates the episode. Successfully navigating to the goal brings $+10$ reward, and correctly crossing the traffic light is awarded with $+10$. Episodes are limited to $200$ time steps. 

Unlike the taxi environment, different situations in this environment require different subsets of features. For example, consider the situation where the agent is at the traffic light, behind another vehicle (Figure \ref{gridworld-traffic}). If the traffic light is red, the agent only needs to rely on this feature to know that it needs to stay in place. If the light is green, the agent needs to observe both the light and the position of the car in front. Under the assumption that the car in front of the agent follows traffic rules, it is likely that the action of the vehicle in front of the agent and agent's action will be highly correlated. However, if the agent relies on this correlation, and only observes the action of the car in front of it to decide whether it should move forward, then its policy will fail in a situation where the car in front stops behaving as expected (e.g., stops even though the traffic light is green or drives despite the red light). This means that the subset of necessary features is not constant throughout the episode, as the action of the vehicle in front needs to be considered in specific states, but can be confusing in other.

\begin{table}[t]
    \centering
    \caption{Training parameters for policies $\pi_{correct}$ and $\pi_{confused}$ in taxi and minigrid environments. The same parameters are used for $\pi_{correct}$ and $\pi_{confused}$ within one task.}
    \begin{adjustbox}{width=1\linewidth}
        \begin{tabular}{@{}ccccccccc@{}} \toprule
            Task & Algorithm &  Architecture & Learning Rate & Gamma & Memory Capacity & 
            \multicolumn{3}{c}{Exploration} \\ 
            & & & & & & Start Epsilon & End Epsilon & Epsilon Decay\\ \cmidrule(r){1-9}
            
             Taxi & DQN & \makecell[c]{Linear(5, 256) \\ Linear(256, 6)} & $1\mathrm{e}{-4}$ & 0.99 & 10000 & 0.9 & 0.01 & 100000 \\ 
            Minigrid traffic & DQN & \makecell[c]{Linear(6, 512) \\ Linear(512, 2)} & $1\mathrm{e}{-4}$ & 0.99 & 80000 & 0.9 & 0.1 & 50000\\
        \bottomrule
        \end{tabular}
    \end{adjustbox}
    \label{hyperparam-policy-simple}
    \vspace{-5pt}
\end{table}

\begin{table}[t]
    \centering
    \caption{Training parameters for feature-parametrized policy $\pi_G$ in taxi and minigrid environments as a part of ReCCoVER algorithm. }
    \begin{adjustbox}{width=1\linewidth}
        \begin{tabular}{@{}ccccccccc@{}} \toprule
            Task & Algorithm & Architecture & Learning Rate & Gamma & Memory Capacity &  \multicolumn{3}{c}{Exploration} \\ 
            & & & & & & Start Epsilon & End Epsilon & Epsilon Decay\\ \cmidrule(r){1-9}
            Taxi & DQN & \makecell[c]{Linear(10, 512) \\ Linear(512, 6)} & $1\mathrm{e}{-4}$ & 0.99 & 320000 & 0.9 & 0.01 & 100000 \\ 
            Minigrid traffic & DQN & \makecell[c]{Linear(12, 512) \\ Linear(512, 2)} & $1\mathrm{e}{-4}$ & 0.99 & 80000 & 0.9 & 0.1 & 50000\\
        \bottomrule
        \end{tabular}
    \end{adjustbox}
    \label{hyperparams-graph-param}
    \vspace{-5pt}
\end{table}

\section{Evaluating ReCCoVER}
\label{Experiments}

To demonstrate detection and correction of causal confusion in RL using ReCCoVER, we set up three evaluation goals to examine:

\begin{enumerate}
    \item \textbf{Goal 1 (Critical states):} Verify that extracted critical states correspond to local maxima states in terms of state-value function.
    \item \textbf{Goal 2 (Recognizing causal confusion):} Verify that, for a policy which relies on a spurious correlation to make a decision in a critical state, ReCCoVER flags the state.
    \item \textbf{Goal 3 (Proposing correct feature subset):} Verify that relying on the feature subset proposed by ReCCoVER during training produces a policy less prone to causal confusion.
\end{enumerate}

To verify that correct critical states have been extracted, we manually examine them, with a view to automate this process in the future. In the taxi environment, local maxima states in terms of the state-value function should correspond to situations in which the agent reaches the passenger location or destination and should perform a pick-up or drop-off action, to complete a sub-goal and collect a reward. In the minigrid traffic environment, local maximum is reached when the agent is at the traffic light, or a step away from the goal, as successfully crossing the traffic light or reaching the goal brings large rewards.

To evaluate the second goal, we train a policy $\pi_{confused}$ which relies on spurious correlations in the environment, and apply ReCCoVER to detect causal confusion in critical states. To train $\pi_{confused}$ in the taxi environment, we purposefully randomize the destination feature during training, forcing $\pi_{confused}$ to rely on the passenger descriptor. As long as the passenger descriptor correlates with the destination, the agent can safely ignore the destination and use the descriptor as its proxy. However, we expect to see a drop in performance in alternative environments in which the destination is intervened on, breaking the descriptor and destination correlation. Additionally, we expect causal confusion to occur only in states in which the destination is important for the policy, such as situations where an agent has picked up the passenger and is trying to drop them off. In the minigrid traffic environment, we randomize the feature denoting the traffic light, in order to learn a policy $\pi_{confused}$ which relies solely on the actions of the vehicle in front of the agent. As long as the vehicle in front of the agent follows the traffic rules (i.e., correctly stops at the traffic light), an agent can safely ignore the traffic light and follow the vehicle. We expect to see causal confusion occur in a situations where vehicle in front does not follow traffic rules.

To evaluate the third goal, we start by examining the alternative environments in which ReCCoVER detected spurious correlations. We manually extract a set of states $S^*$ where causal confusion damaged the performance of $\pi_{confused}$ and propose a different feature subset $\mathcal{F}(S^*)$ to be used in states in $S^*$, to prevent learning spurious correlations. $\mathcal{F}(S^*)$ equals the feature subset $G'$ of policy $\pi_G(G')$ which outperformed $\pi_{confused}$ in alternative environments. We then train a new policy $\pi_{correct}$, ensuring that it focuses only on features from $\mathcal{F}(S^*)$ in states in $S^*$. We do this by purposefully randomizing all features not present in $\mathcal{F}(S^*)$, in states from $S^*$ during training. Finally, we apply ReCCoVER to $\pi_{correct}$, to examine whether adjusting which features policy relies on in certain areas of the state space makes it less prone to learning spurious correlations. 

Training parameters for policies $\pi_{correct}$ and $\pi_{confused}$ in both environments are shown in Table \ref{hyperparam-policy-simple}. ReCCoVER also requires parameters $k$ and $\alpha$, denoting the number of steps that policies are evaluated in alternative environments and novelty sensitivity when choosing novel alternative environments respectively. We evaluate policies in alternative environments for $k=3$ steps in taxi, and $k=1$ steps in minigrid traffic environment, and use $\alpha = 0.9$ in both tasks. Additionally, training parameters for feature-parametrized policy $\pi_G$ in ReCCoVER are given in Table \ref{hyperparams-graph-param}. 
We limit the training of $\pi_G$ by including in $\mathcal{G}$ only those feature subsets which enable learning at least a part of the task. In the taxi environment, agent's state consists of its x location, y location, passenger descriptor, passenger location and destination. We consider $4$ feature subsets corresponding to binary vectors $\begin{bmatrix} 1 & 1 & 1 & 1 & 1 \end{bmatrix}$, $\begin{bmatrix} 1 & 1 & 0 & 1 & 1 \end{bmatrix}$, $\begin{bmatrix} 1 & 1 & 1 & 1 & 0 \end{bmatrix}$ and $\begin{bmatrix} 1 & 1 & 0 & 1 & 0 \end{bmatrix}$. First three feature subsets can be used to learn the complete task in the environment where descriptor and destination features are highly correlate, as they contain all necessary information. The last subset can help the agent achieve the first subtask of picking up the passenger, but lacks information about destination, necessary for completing the task. In minigrid traffic environment, agent's state consists of its location, goal location, other vehicle's action and location, traffic light location and color. We limit $\mathcal{G}$ to $3$ feature subsets: $\begin{bmatrix} 1 & 1 & 1 & 1 & 1 & 1 \end{bmatrix}$, $\begin{bmatrix} 1 & 1 & 1 & 1 & 1  & 0\end{bmatrix}$ and $\begin{bmatrix} 1 & 1 & 0 & 1 & 1  & 1\end{bmatrix}$. All three subsets can learn the task fully in the environment in which the vehicle in front of the agent is following traffic rules. 

\section{Results}

In this section, we present our results of evaluating ReCCoVER in two RL environments against the three evaluation goals (Section \ref{Experiments}).

\subsection{Taxi Environment}

\subsubsection*{Goal 1 (Extracting Critical States):} ReCCoVER extracts 16 critical states (Table \ref{output-taxi}). $S_c$ contains states in which the agent is at the passenger location and should perform the pick up action, and states in which the agent should perform a drop off action, after successfully picking up the passenger and reaching the destination. Both situations correspond to local maxima, as they represent states in which agent has potential to receive a large reward in the next step. 

\subsubsection*{Goal 2 (Recognizing Causal Confusion):} We apply ReCCoVER to examine the $\pi_{confused}$, policy which relies on the passenger descriptor, but ignores the destination feature. For each critical state, ReCCoVER generated on average $8.81$ alternative environments (Table \ref{output-taxi}), and examined behavior of $\pi_{confused}$ and $\pi_G$ in them. ReCCoVER detected causal confusion in $4$ critical states (Table \ref{output-taxi}), corresponding to $4$ possible destination positions. In these states, the agent is at the destination, about to drop off the passenger. Intervening on the destination feature in these states breaks the correlation between descriptor and destination, and leads to poor performance for $\pi_{confused}$. For that reason, we discover that performance of $\pi_{confused}$ drops in alternative environments obtained by intervening on the destination feature. However, in the same alternative environments, policy $\pi_G(\begin{bmatrix} 1 &  1 & 0 &  1 &  1 \end{bmatrix})$ performed significantly better than $\pi_{confused}$. Policy $\pi_G(\begin{bmatrix} 1 &  1 &  0 &  1 &  1 \end{bmatrix})$ relied on destination and ignored the passenger descriptor, making it robust to changes in correlation between the two features. In critical states where the agent should pick up the passenger, ReCCoVER does not detect causal confusion, because information about destination is not necessary for this part of the task and ignoring it does not damage immediate performance. 
    
\subsubsection*{Goal 3 (Proposing Correct Feature Subset):} All alternative environments in which ReCCoVER detected causal confusion for policy $\pi_{confused}$ have been obtained by intervening on destination, in situations where the agent already picked up the passenger. For this reason, we propose that the subset of features that the agent relies on be altered only in parts of the state space in which the agent has completed the subtask of picking up the passenger. Policy $\pi_{confused}$ was outperformed in each such alternative environment by $\pi_G(\begin{bmatrix} 1 &  1 &  0 &  1 &  1 \end{bmatrix})$. For this reason, train $\pi_{correct}$ to rely on feature subset $\begin{bmatrix} 1 &  1 &  0 &  1 &  1 \end{bmatrix}$ in parts of the state space where the passenger has been picked up. In other areas of the environment, $\pi_{correct}$ relies on the same features as $\pi_{confused}$, denoted by $\begin{bmatrix} 1 &  1 &  1 &  1 &  0 \end{bmatrix}$. As above, we randomize the features that the agent should not rely on in specific states during training. Finally, we apply ReCCoVER algorithm to uncover whether altered feature subset helped $\pi_{correct}$ become less susceptible to causal confusion. ReCCoVER does not detect causal confusion in any critical state for policy $\pi_{correct}$, indicating that it does not rely on spurious correlations (Table \ref{output-taxi}).

\begin{table}[t]
    \caption{Output of each stage of the ReCCoVER algorithm in taxi environment.}
    \centering
    \begin{adjustbox}{width=\textwidth}
        \begin{tabular}{@{}cccccc@{}}
        \toprule
         Policy & \makecell[c]{Number of \\episodes} & \makecell[c]{Number of \\ collected transitions} &  \makecell[c]{Number of \\critical states} & \makecell[c]{Average number of alternative \\worlds per critical state} & \makecell[c]{Number of states\\ where causal confusion detected}\\ \midrule
         
         \makecell[c]{$\pi_{confused}$ \\ $\pi_{correct}$} & \makecell[c]{100\\100} & \makecell[c]{920 \\ 916} & \makecell[c]{16 \\ 16} & \makecell[c]{8.81 \\ 8.43} & \makecell[c]{4 \\ 0} \\
        
        \bottomrule
        \end{tabular}
    \end{adjustbox}
    \label{output-taxi}
\end{table}

\subsection{Minigrid Traffic Environment}

\subsubsection*{Goal 1 (Extracting Critical States):} ReCCoVER extracts $2$ critical states (Table \ref{output-minigrid}). The first represents the situation where the agent is at the traffic light, while in the second, the agent is one step away from the goal. Both states show to situations in which the agent is about to complete a sub task or reach the goal and receive a large reward, making them plausable local maxima states.

\subsubsection*{Goal 2 (Recognizing Causal Confusion):} We apply ReCCoVER to examine behavior of policy $\pi_{confused}$, which fully relies on the actions of the vehicle in front of the agent to make decisions, ignoring the traffic light. On average, $14$ alternative environments are generated for each critical state (Table \ref{output-minigrid}). ReCCoVER detects causal confusion in alternative environment where the agent is at the red light, but the vehicle in front of it continues driving (Table \ref{output-minigrid}). Policy $\pi_{confused}$, relying only on the decisions of the vehicle in front, in this situation runs the red light, and receives a large penalty. However, policy $\pi_G(\begin{bmatrix} 1 & 1 &  0 &  1 &  1 & 1 & 1 \end{bmatrix})$, which relies on the traffic light, but ignores the actions of the vehicle in front, acts correctly in this state and stops at the red light. It is important to note that $\pi_G(\begin{bmatrix} 1 & 1 &  0 &  1 &  1 & 1 & 1 \end{bmatrix})$ cannot be applied to entire task, as knowing the current action of the vehicle in front is needed to avoid collision. In a critical state where the agent is one step away from the goal, no causal confusion is detected, since the traffic light in this state does not affect the action. 
    
\subsubsection*{Goal 3 (Proposing Correct Feature Subset):} By examining the alternative environment in which ReCCoVER detected causal confusion for policy $\pi_{confused}$, we propose that the feature subset that the agent relies on be altered in part of the state space $S^*$ where the agent is at the red traffic light. We train policy $\pi_{correct}$ to rely on the same feature subset $\begin{bmatrix} 1 &  1 &  1 &  1 & 1 &  0 \end{bmatrix}$ as $\pi_{confused}$ in all situations except in states from $S^*$, where it should rely on subset $\begin{bmatrix} 1 & 1 &  0  &  1 & 1 & 1 \end{bmatrix}$. We train $\pi_{correct}$ to ignore the traffic light when agent is not directly at it, by randomizing the traffic light feature. However, when the agent is at the red light, we encourage the agent to observe the traffic light and ignore the vehicle in front, by randomizing the feature corresponding to the action of the vehicle. This way, we obtain a policy $\pi_{correct}$, which focuses on the car in front of it in all situations, except when it faces a red light, where it ignores the vehicle and relies only on the light color. To ensure that $\pi_{correct}$ can encounter all the necessary situations, we train it in an environment where the vehicle in front does not always follow the traffic rules, and might run the red light. ReCCoVER is applied to examine the behavior of $\pi_{correct}$, and causal confusion is not found in any critical states, indicating that $\pi_{correct}$ does not rely on any spurious correlations (Table \ref{output-minigrid}).

\begin{table}[t]
    \caption{Output of each stage of the ReCCoVER approach in minigrid environment.}
    \centering
    \begin{adjustbox}{width=\textwidth}
        \begin{tabular}{@{}cccccc@{}}
        \toprule
        Policy & \makecell[c]{Number of \\episodes} & \makecell[c]{Number of \\ collected transitions} &  \makecell[c]{Number of \\critical states} & \makecell[c]{Average number of alternative \\worlds per critical state} & \makecell[c]{Number of states\\ where causal confusion detected}\\ \midrule
        
        \makecell[c]{$\pi_{confused}$ \\$\pi_{correct}$} & \makecell[c]{100\\100} & \makecell[c]{1500 \\ 1500} & \makecell[c]{2 \\ 1} & \makecell[c]{14 \\ 15} & \makecell[c]{1\\ 0} \\
        
        \bottomrule
        \end{tabular}
    \end{adjustbox}
    \label{output-minigrid}
\end{table}

\section{Discussion and future work}

In this work, we explored the problem of causal confusion in RL and proposed ReCCoVER, a method for detecting spurious correlations in agent's reasoning before deployment. We evaluated our approach in two RL environments, where ReCCoVER located situations in which reliance on spurious correlations was damaging agent's performance, and proposed a different subset of features that should instead be used in that area of state space. Our work is, however, of exploratory nature, and its purpose was to perform initial analysis of if and how detecting causal confusion can aid in developing more transparent RL agents. As such, certain stages of ReCCoVER algorithm are simplified and manual -- for example, we examine agent's behavior only in critical states, and limit the search for causal confusion to a subset of alternative environments. 

For ReCCoVER to be applicable beyond benchmark scenarios, it needs to be extended in two main directions -- increasing and verifying its scalability and automating the verification of ReCCoVER. In future work, we hope to explore ways for narrowing down the feature subset search space $\mathcal{G}$, while training the feature-parametrized policy, to make this stage of ReCCoVER more feasible. Additionally, we hope to automate parts of the process which currently require manual attention, such as verification of critical states or extraction of areas of the state space where causal confusion is damaging. Once fully automated, we will explore applicability of our approach beyond explainability, due to its focus on correcting mistakes in agent's reasoning. Mistake detection and correction is useful in transfer learning, to ensure that knowledge that is being reused is not prone to causal confusion. Additionally, it can help improve robustness of RL policies, either through human-in-the-loop approaches or as auto-correction. 

\section*{Acknowledgement}

This publication has emanated from research conducted with the financial support of a grant from Science Foundation Ireland under Grant number 18/CRT/6223. For the purpose of Open Access, the author has applied a CC BY public copyright licence to any Author Accepted Manuscript version arising from this submission.

%
%
%
\bibliographystyle{splncs04}
\bibliography{references}

\begin{thebibliography}{10}
\providecommand{\url}[1]{\texttt{#1}}
\providecommand{\urlprefix}{URL }
\providecommand{\doi}[1]{https://doi.org/#1}

\bibitem{amir2018highlights}
Amir, D., Amir, O.: Highlights: Summarizing agent behavior to people. In:
  Proceedings of the 17th International Conference on Autonomous Agents and
  MultiAgent Systems. pp. 1168--1176 (2018)

\bibitem{brockman2016openai}
Brockman, G., Cheung, V., Pettersson, L., Schneider, J., Schulman, J., Tang,
  J., Zaremba, W.: Openai gym. arXiv preprint arXiv:1606.01540  (2016)

\bibitem{burkart2021survey}
Burkart, N., Huber, M.F.: A survey on the explainability of supervised machine
  learning. Journal of Artificial Intelligence Research  \textbf{70},  245--317
  (2021)

\bibitem{carvalho2019machine}
Carvalho, D.V., Pereira, E.M., Cardoso, J.S.: Machine learning
  interpretability: A survey on methods and metrics. Electronics
  \textbf{8}(8), ~832 (2019)

\bibitem{gym_minigrid}
Chevalier-Boisvert, M., Willems, L., Pal, S.: Minimalistic gridworld
  environment for openai gym. \url{https://github.com/maximecb/gym-minigrid}
  (2018)

\bibitem{coppens2019distilling}
Coppens, Y., Efthymiadis, K., Lenaerts, T., Now{\'e}, A., Miller, T., Weber,
  R., Magazzeni, D.: Distilling deep reinforcement learning policies in soft
  decision trees. In: Proceedings of the IJCAI 2019 workshop on explainable
  artificial intelligence. pp.~1--6 (2019)

\bibitem{dethise2019cracking}
Dethise, A., Canini, M., Kandula, S.: Cracking open the black box: What
  observations can tell us about reinforcement learning agents. In: Proceedings
  of the 2019 Workshop on Network Meets AI \& ML. pp. 29--36 (2019)

\bibitem{gajcin2021contrastive}
Gajcin, J., Nair, R., Pedapati, T., Marinescu, R., Daly, E., Dusparic, I.:
  Contrastive explanations for comparing preferences of reinforcement learning
  agents. arXiv preprint arXiv:2112.09462  (2021)

\bibitem{greydanus2018visualizing}
Greydanus, S., Koul, A., Dodge, J., Fern, A.: Visualizing and understanding
  atari agents. In: International conference on machine learning. pp.
  1792--1801. PMLR (2018)

\bibitem{de2019causal}
de~Haan, P., Jayaraman, D., Levine, S.: Causal confusion in imitation learning.
  Advances in Neural Information Processing Systems  \textbf{32},  11698--11709
  (2019)

\bibitem{lundberg2017unified}
Lundberg, S.M., Lee, S.I.: A unified approach to interpreting model
  predictions. Advances in neural information processing systems  \textbf{30}
  (2017)

\bibitem{lyle2021resolving}
Lyle, C., Zhang, A., Jiang, M., Pineau, J., Gal, Y.: Resolving causal confusion
  in reinforcement learning via robust exploration. In: Self-Supervision for
  Reinforcement Learning Workshop-ICLR 2021 (2021)

\bibitem{madumal2020explainable}
Madumal, P., Miller, T., Sonenberg, L., Vetere, F.: Explainable reinforcement
  learning through a causal lens. In: Proceedings of the AAAI conference on
  artificial intelligence. vol.~34, pp. 2493--2500 (2020)

\bibitem{mnih2013playing}
Mnih, V., Kavukcuoglu, K., Silver, D., Graves, A., Antonoglou, I., Wierstra,
  D., Riedmiller, M.: Playing atari with deep reinforcement learning. arXiv
  preprint arXiv:1312.5602  (2013)

\bibitem{olson2019counterfactual}
Olson, M.L., Neal, L., Li, F., Wong, W.K.: Counterfactual states for atari
  agents via generative deep learning. arXiv preprint arXiv:1909.12969  (2019)

\bibitem{scm}
Pearl, J.: Causality. Cambridge university press (2009)

\bibitem{peters2017elements}
Peters, J., Janzing, D., Sch{\"o}lkopf, B.: Elements of causal inference:
  foundations and learning algorithms. The MIT Press (2017)

\bibitem{puiutta2020explainable}
Puiutta, E., Veith, E.: Explainable reinforcement learning: A survey. In:
  International cross-domain conference for machine learning and knowledge
  extraction. pp. 77--95. Springer (2020)

\bibitem{puri2019explain}
Puri, N., Verma, S., Gupta, P., Kayastha, D., Deshmukh, S., Krishnamurthy, B.,
  Singh, S.: Explain your move: Understanding agent actions using specific and
  relevant feature attribution. arXiv preprint arXiv:1912.12191  (2019)

\bibitem{sequeira2020interestingness}
Sequeira, P., Gervasio, M.: Interestingness elements for explainable
  reinforcement learning: Understanding agents' capabilities and limitations.
  Artificial Intelligence  \textbf{288},  103367 (2020)

\bibitem{simonyan2013deep}
Simonyan, K., Vedaldi, A., Zisserman, A.: Deep inside convolutional networks:
  Visualising image classification models and saliency maps. arXiv preprint
  arXiv:1312.6034  (2013)

\bibitem{csimcsek2004using}
{\c{S}}im{\c{s}}ek, {\"O}., Barto, A.G.: Using relative novelty to identify
  useful temporal abstractions in reinforcement learning. In: Proceedings of
  the twenty-first international conference on Machine learning. p.~95 (2004)

\bibitem{verma2018programmatically}
Verma, A., Murali, V., Singh, R., Kohli, P., Chaudhuri, S.: Programmatically
  interpretable reinforcement learning. In: International Conference on Machine
  Learning. pp. 5045--5054. PMLR (2018)

\bibitem{van2018contrastive}
van~der Waa, J., van Diggelen, J., Bosch, K.v.d., Neerincx, M.: Contrastive
  explanations for reinforcement learning in terms of expected consequences.
  arXiv preprint arXiv:1807.08706  (2018)

\end{thebibliography}

\end{document}